# Multi-task Domain Adaptation for Sequence Tagging

**Nanyun Peng** and **Mark Dredze**
Human Language Technology Center of Excellence
Center for Language and Speech Processing
Johns Hopkins University, Baltimore, MD, 21218
`npeng1@jhu.edu, mdredze@cs.jhu.edu`

## Abstract

Many domain adaptation approaches rely on learning cross domain shared representations to transfer the knowledge learned in one domain to other domains. Traditional domain adaptation only considers adapting for one task. In this paper, we explore multi-task representation learning under the domain adaptation scenario. We propose a neural network framework that supports domain adaptation for multiple tasks simultaneously, and learns shared representations that better generalize for domain adaptation. We apply the proposed framework to domain adaptation for sequence tagging problems considering two tasks: Chinese word segmentation and named entity recognition. Experiments show that multi-task domain adaptation works better than disjoint domain adaptation for each task, and achieves the state-of-the-art results for both tasks in the social media domain.

## 1 Introduction

Many natural language processing tasks have abundant annotations in formal domain (news articles) but suffer a significant performance drop when applied to a new domain, where only a small number of annotated examples are available. The idea behind domain adaptation is to leverage annotations from high-resource (source) domains to improve predictions in low-resource (target) domains by training a predictor for a single task across different domains.

Domain adaptation work tends to focus on changes in data distributions, e.g. different words are used in each domain. Domain adaptation methods include unsupervised (Blitzer et al., 2006) and supervised (Daumé III, 2007) variants, depending on whether there exists no or some training data in the target domain. This paper considers the case of supervised domain adaptation, where we have a limited amount of target domain training data, but much more training data in a source domain.

Work on domain adaptation mostly follows two approaches: parameter tying (i.e. linking similar features during learning) (Dredze and Crammer, 2008; Daumé III, 2007, 2009; Finkel and Manning, 2009; Kumar et al., 2010; Dredze et al., 2010), and learning cross domain representations (Blitzer et al., 2006, 2007; Glorot et al., 2011; Chen et al., 2012; Yang and Eisenstein, 2015). Often times, domain adaptation is formulated as learning a single model for the same task across domains, although with a focus on maximizing target domain performance. This is similar in spirit to multi-task learning (MTL) (Caruana, 1997) which jointly learns models for several tasks, for example. learning a single data representation common to each task (Ando and Zhang, 2005; Collobert et al., 2011; Liu et al., 2016c; Peng and Dredze, 2016; Yang et al., 2016; Liu et al., 2016a). Given the similarity between domain adaptation and MTL, it is natural to ask: can domain adaptation benefit from jointly learning across several tasks?

This paper investigates how MTL can induce better representations for domain adaptation. There are several benefits. First, learning multiple tasks provides more training data for learning. Second, MTL provides a better inductive learning bias so that the learned representations better generalize. Third, considering several tasks in domain adaptation opens up the opportunities to adapt from a different domain *and* a different task, a mismatch setting which has not previously been explored. We present a representation learning framework based on MTL that incorporates parameter tying strategies common in domain adaptation. Our framework is based on a bidirectional long short-term memory network with a conditional random fields (BiLSTM-CRFs) (Lample et al., 2016) for sequence tagging. We consider sequence tagging problem since they are common in NLP applications and have been demonstrated to benefit from learning representations (Lample et al., 2016; Yang et al., 2016; Peng and Dredze, 2016; Ma and Hovy, 2016).

This paper makes the following contributions:

- A neural MTL domain adaptation framework that considers several tasks *simultaneously* when doing domain adaptation.

- A new domain/task mismatch setting: where you have two datasets from two different, but related domains and tasks.

- State-of-the-art results on Chinese word segmentation and named entity recognition in social media data.

## 2 Model

We begin with a brief overview of our model, and then instantiate each layer with specific neural architectures to conduct multi-task domain adaptation for sequence tagging. Figure 1 summarizes the entire model presented in this section.

A representation learner that is shared across all domains and tasks, and learns robust data representations for features. This feeds a domain projection layer, with one projection for each domain that transforms the learned representations for different domains into the same shared space. As a result, the final layer of task specific models, which learns feature weights for different tasks, can be shared across domains since the learned representations (features) for different domains are now in the same space. The framework is flexible in both the number of tasks and domains. Increasing the number of domains linearly increases domain projection parameters, with the number of other model parameters unchanged. Similarly, increasing the number of tasks only linearly increases the number of task specific model parameters. If there is only one domain, then the framework reduces to a multi-task learning framework, and similarly, the framework reduces to a standard domain adaptation framework if there is only one task.

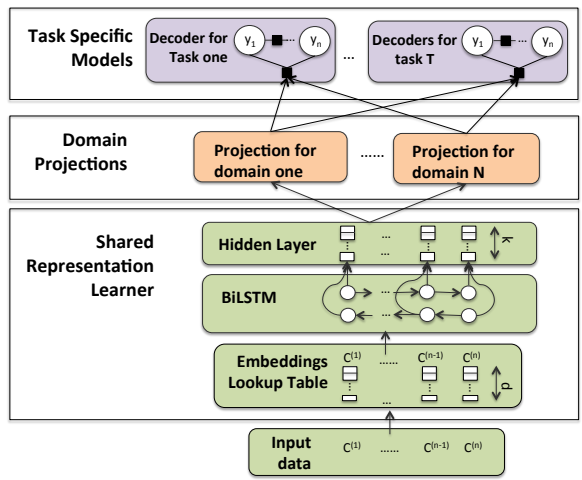

Figure 1: An overview of our proposed model framework. The bottom layer is shared by all tasks and domains. The domain projections contain one projection per domain and the task specific models (top layer) contain one model per task.

The shared representation learner, domain projections and task specific models can be instantiated based on the application. In this paper, we focus on sequence tagging problems. We now introduce our instantiated neural architecture for multi-task domain adaptation for sequence tagging.

### 2.1 BiLSTM for representation learning

Long short-term memory (LSTM) (Hochreiter and Schmidhuber, 1997) is a type of recurrent neural network (RNN) that models interdependencies in sequential data. It addresses the vanishing or exploding gradients (Bengio et al., 1994; Pascanu et al., 2013) problems of vanilla RNNs by using a series of gates (input, forget and output gates) to control how memory is propagated in the hidden states of the model, and thus effectively captures long-distance dependencies between the inputs.

Many NLP applications use bi-directional LSTMs (BiLSTM) (Dyer et al., 2015) to scan both left-to-right and right-to-left, which capture left and right context. The hidden vectors produced by both LSTMs are concatenated to form the final output vector $h_t = \overrightarrow{h_t} \oplus \overleftarrow{h_t}$. BiLSTMs have become a common building block for learning representations in NLP and have achieved impressive performance in problems such as sequence tagging (Lample et al., 2016; Yang et al., 2016; Ma and Hovy, 2016), relation classification (Xu et al., 2015; Zhang et al., 2015), and syntactic parsing (Kiperwasser and Goldberg, 2016; Cross and Huang, 2016). We use a BiLSTM as our representation learner. It produces a hidden vector for each token in the sentence, which we denote as:

$$h_t = \text{BiLSTM}(x_{1:n}, t) \quad (1)$$

where $x_{1:n}$ denotes the whole input sequence of length $n$, and $t$ denotes the $t$-th position. The representation for the whole sequence is thus denoted as $\boldsymbol{h} = h_{1:n}$.

## 2.2 Domain Projections

Domain adaptation requires learning a shared representation that generalizes across domains. Ideally, parameter estimation of the BiLSTM should learn to produce such robust features. However, this may place a heavy burden on the BiLSTM; it does not know the identity of each domain yet must still learns how to map two heterogeneous input types to the same representation. To reduce this burden, we introduce a domain projection layer, which relies on explicit domain specific transformation functions to produce shared representations. We place this transformation between the representation learner and the task specific predictor to alleviates pressure on the representation learner to learn cross domain representations. Note that the domain projection layer works *jointly* with the representation learner to produce shared representations. We experiment with two simple strategies for domain projections which are based on previous lines of work in domain adaptation.

### 2.2.1 Domain Masks

The first strategy is inspired by Daumé III (2007) and Yang and Hospedales (2015), which split the representations into several regions, with one region shared among domains, and others specific for each domain. As a result, the BiLSTM representation learner will learn to put the features that are suitable to be shared across domains into the shared region, and domain specific features to the corresponding region for the domain.

We implement this strategy by defining domain masks $\boldsymbol{m}_d$, which is a vector for the $d$th domain. The mask $\boldsymbol{m}_d$ has value 1 for the effective dimensions of domain $d$ and domain shared region, and 0 for all other dimensions. For example, assume we have two domains and a $k$ dimensional hidden vector for features, the first $k/3$-dimensions is shared between the two domains, while the $k/3+1$ to $2k/3$ dimensions are used only for domain 1, and the remaining dimensions for domain 2. The mask for domain 1 and domain 2 would be:

$$\boldsymbol{m}_1 = [\vec{1}, \vec{1}, \vec{0}], \quad \boldsymbol{m}_2 = [\vec{1}, \vec{0}, \vec{1}]. \quad (2)$$

We can then apply these masks directly to the hidden vectors $\boldsymbol{h}$ learned by the BiLSTM to produce a projected hidden state $\hat{\boldsymbol{h}}$:

$$\hat{\boldsymbol{h}} = \boldsymbol{m}_d \odot \boldsymbol{h}, \quad (3)$$

where $\odot$ denotes element-wise multiplication. Since only a subset of the dimensions are used as features in each domain, the BiLSTM will be encouraged to learn to partition the dimensions of the output hidden vectors into domains.

Note that in Daumé III (2007), the domain masks operate on hand engineered features, thus only affect feature weights. However, here the domain masks will change the parameters learned in BiLSTMs as well, changing the learned features. Therefore, training data from one domain will also change the other domains' representation. When we jointly train with data from all domains, the model has to balance the training objectives for all domains simultaneously.

### 2.2.2 Linear Projection

The second domain adaptation strategy we explore is a linear transformation to each domain, denoted as $T_d$. Given a $k$-dimensional vector representation $\boldsymbol{h}$, $T_d$ is a $k \times k$ matrix that projects the learned BiLSTM hidden vector to a common space that can be used by a shared task specific model. We use the transformation:

$$\hat{\boldsymbol{h}} = T_d \boldsymbol{h}. \quad (4)$$

We learn $T_d$ for each domain jointly with other model parameters. While this model has greater freedom in learning representations across domains, it relies on the training data to learn a good transformation, and does not explicitly partition the representations into domain regions.

## 2.3 Task Specific Neural-CRF Models

Multi-task domain adaptation *simultaneously* considers several tasks adapting domains since the related tasks would help induce more robust data representations for domain adaptation. Additionally, it enables leveraging more data to learn better domain projections. The goal of a task specific model is to learn parameters to project the shared representations to the desired outputs for the corresponding task. Different tasks that define different output spaces need separate task specific models.

For our applications to sequence tagging problems, we choose Conditional Random Fields (CRFs) (Lafferty et al., 2001) as task specific models, since it is widely used in previous work and is shown to benefit from learning representations (Peng and Dredze, 2015; Lample et al., 2016; Ma and Hovy, 2016). These "Neural-CRFs" define the conditional probability of a sequence of labels given the input as:

$$p(\boldsymbol{y}^k|\boldsymbol{x}^k;W) = \frac{\prod_{i=1}^n \exp\left(W^T F(y_{i-1}^k, y_i^k, \psi(\boldsymbol{x}^k))\right)}{Z^k},$$

where $i$ indexes the position in the sequence, $F$ is the feature function, and $\psi(\boldsymbol{x}^k)$ defines a transformation of the original input, in our case $\psi(\boldsymbol{x}^k) = BiLSTM(\boldsymbol{x}^k)$. $Z^k$ is the partition function defined as:

$$Z^k = \sum_{\boldsymbol{y} \in \mathcal{Y}} \prod_{i=1}^n \exp\left(W^T F(y_{i-1}^k, y_i^k, \psi(\boldsymbol{x}^k))\right).$$

### 2.3.1 Sharing Task Specific Models

We could create a CRF decoder for each task and domain. This is the practice of some (Yang and Hospedales, 2015) who consider domain adaptation, or multi-domain learning, a special case of MTL, and learn separate models for the same task from different domains.

Instead, we argue that learning a single model for a task regardless of the number of domains draws strong connections to the traditional domain adaptation literature. It enjoys the benefit of increasing the amount of training data for each task by considering different domains, and better handles the problem of shifts in data distributions by explicitly considering different domains. Therefore, we use a single CRF per *task*, shared across all domains.

## 3 Parameter Estimation

The proposed neural architecture for multi-task domain adaptation can be trained end-to-end by maximizing data log-likelihood. As there are $D \times T$ [1] datasets, the final loss function is a linear combination of the log-likelihood of each dataset. For simplicity, we give each dataset equal weight when forming the linear combination.

---
[1] $D$ denotes the number of domains and $T$ the number of tasks

**Training** Model training is a straightforward application of gradient based back-propagation. We use alternating optimization among each dataset with stochastic gradient descent (SGD). To prevent training from skewing the model to a specific dataset due to the optimization order, we subsample the number of instances used in each epoch with a fraction $\lambda$ w.r.t. the smallest dataset size, which is tuned as a hyper-parameter on development data. A separate learning rate is tuned for each dataset, and we decay the learning rate when results on development data do not improve after 5 consecutive epochs. We train for up to 30 epochs and use early stopping (Caruana et al., 2001; Graves et al., 2013) as measured on development data. We select the best model for each dataset based on hyper-parameter tuning. We use dropout on the embeddings and the BiLSTM output vectors as in Ma and Hovy (2016).

**Initialization** We use pre-trained Chinese embeddings provided by Peng and Dredze (2015) with dimension 100. All other model parameters are initialized uniformly at random in the range of $[-1, 1]$.

**Inference** For training the CRFs, we use marginal inference and maximize the marginal probabilities of the labels in the training data. At test time, the label sequence with highest conditional probability $y^* = \arg\max p(y|x; \Omega)$ is obtained by MAP inference.

**Hyper-parameters** Our hyper-parameters include the initial learning rate (per dataset, in the range of [0.005, 0.01, 0.02]), the dropout rate for the input embedding and the hidden vectors (in the range of [0, 0.1, 0.2]), and the subsample coefficient for each setting (in the range of [5, 10, 15]). We tune these hyper-parameter using beam search on development data. For convenience, the embedding and the LSTM hidden vector dimensions are set to 100 and 150 respectively.

## 4 Experimental Setup

We test the effectiveness of the multi-task domain adaptation framework on two sequence tagging problems: Chinese word segmentation (CWS) and named entity recognition (NER). We consider two domains: news and social media, with news the source domain and social media the target domain.

| Dataset | #Train | #Dev | #Test |
|---|---|---|---|
| **SighanCWS** | 39,567 | 4,396 | 4,278 |
| **SighanNER** | 16,814 | 1,868 | 4,636 |
| **WeiboCWS** | 1,600 | 200 | 200 |
| **WeiboNER** | 1,350 | 270 | 270 |

Table 1: Datasets statistics.

### 4.1 Datasets

We consider two domains: news and social media for the two tasks: CWS and NER. This results in four datasets: news CWS data comes from the SIGHAN 2005 shared task (*SighanCWS*) (Emerson, 2005), news NER data comes from the SIGHAN 2006 shared task (*SighanNER*) (Levow, 2006), social CWS data (*WeiboSeg*) created by Zhang et al. (2013), and social NER data (*WeiboNER*) created by Peng and Dredze (2015).

Both *SighanCWS* and *SighanNER* contain several portions[2]; we use those for simplified Chinese (PKU and MSR respectively). The datasets do not have development data, so we hold out the last 10% of training data for development. *SighanNER* contains three entity types (person, organization and location), while *WeiboNER* is annotated with four entity types (person, organization, location and geo-political entity), including named and nominal mentions. To match the two tag sets, we only use named mentions in *WeiboNER* and merge geo-political entities and locations. The 2000 annotated instances in *WeiboSeg* were meant only for evaluation, so we split the data ourselves using an 8:1:1 split for training, development, and test. Hyper-parameters are tuned on the development data and we report the precision, recall, and F1 score on the test portion. Detailed data statistics is shown in Table 1.

### 4.2 Baselines

We consider two baselines common in domain adaptation experiments. The first baseline only considers a single dataset at a time (*separate*) by training *separate* models just on in-domain training data. The second baseline (*mix*) uses out-of-domain training data for the same task by mixing it with the in-domain data. For both the baselines, we use the BiLSTM-CRFs neural architec-ture (Lample et al., 2016), which achieved state-of-the-art results on NER and other sequence tagging tasks (Peng and Dredze, 2016; Ma and Hovy, 2016; Yang et al., 2016).

[2]The portions are annotated by different institutes, and cover both traditional and simplified Chinese

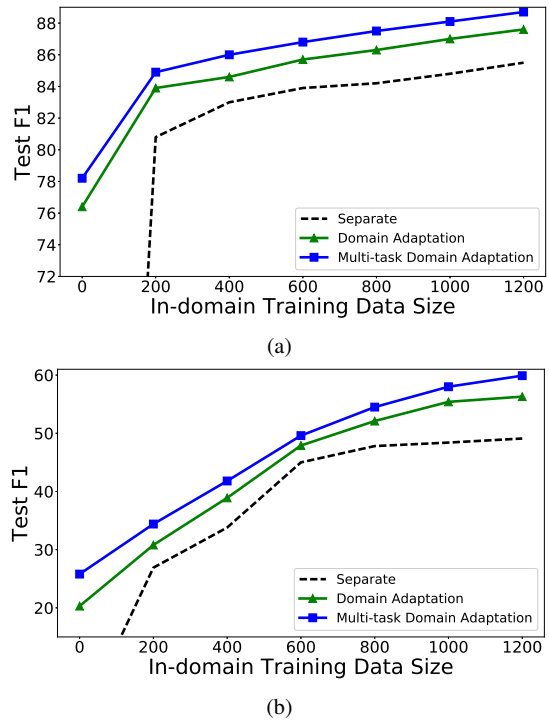

Figure 2: The effect of training data size on social media CWS (top) and NER (bottom) tasks. With more in-domain training data we see diminishing returns from domain adaptation. Our proposed multi-task domain adaptation framework is also applicable for unsupervised domain adaptation (with no in-domain training data).

## 5 Experimental Results

### 5.1 Main Results

Table 2 presents the results for domain adaptation to the target domain (social media) test data. The baseline method *Mix* improves over *Separate* as it benefits from the increased training data. The single task domain adaptation models are a special case of the proposed multi-task domain adaptation framework: with only one task specific model in the top layer (CWS or NER). Both of our approaches (domain mask and linear projection) improve over the baseline methods. Knowing the domain of the training data helps the model better learn effective representations. Finally, we see further improvements in the multi-task domain adaptation setting. By considering additional tasks in addition to domains, we achieve new state-of-the-art results on the two tasks. We compare to the best published results from Zhang et al. (2013) and Peng and Dredze (2016) with F1 scores of 87.5% (CWS) and 55.3% (NER), respectively.

| Settings | Datasets / Methods | CWS Prec | CWS Recall | CWS F1 | NER Prec | NER Recall | NER F1 |
|---|---|---|---|---|---|---|---|
| **Baseline** | Separate | 86.2 | 85.7 | 86.0 | 57.2 | 42.1 | 48.5 |
| | Mix | 87.0 | 86.1 | 86.5 | 60.9 | 44.0 | 51.1 |
| **Domain Adapt** | Domain Mask | 88.7 | 87.1 | 87.9 | 68.2 | 48.6 | 56.8 |
| | Linear Projection | 88.0 | 87.5 | 87.7 | 73.3 | 45.8 | 56.4 |
| **Multi-task DA** | Domain Mask | **89.7** | 88.3 | **89.0** | 60.2 | **52.3** | **59.9** |
| | Linear Projection | 89.1 | **88.6** | 88.9 | **68.6** | 49.5 | 57.5 |

Table 2: Test results for CWS and Chinese NER on the target social media domain. The first two rows are baselines (Section 4.2,) followed by two domain adaptation models that only considers one task a time. The last two rows are the proposed multi-task domain adaptation framework building upon the two domain adaptation models, respectively. Domain adaptation models leverage out-of-domain training data and *significantly* improve over the *Separate* baseline, as well as the *Mix* baseline which trains with the out-of-domain data without considering domain shift. Multi-task domain adaptation further *significantly* improves over traditional domain adaptation on both domain adaptation models and achieved the new state-of-the-art results on the two tasks.

**Statistical Significance** We measures statistical significance using McNemars chi-square test (McNemar, 1947) for paired significant test. We treated the predicted spans (not tokens) that agreed with the ground truth as positive, otherwise negative. For the NER task, we only count the spans that corresponds to named entities. We compare the best baseline (*mix*) and the two domain adaptation models, as well as between the domain adaptation models and their multi-task domain adaptation counterpart. Both the domain adaptation models *significantly* improved over the *mix* baseline ($p < 0.01$), and the multi-task domain adaptation methods *significantly* improved over their single task domain adaptation counterpart ($p < 0.01$). We cannot conduct paired significance tests with the best published results since we do not have access to their outputs.

## 5.2 In-domain Training Data

We also conducted several experiments to show the flexibility of our multi-task domain adaptation framework and analyze the behavior of the models by varying the training data.

We first consider the effect of in-domain training data size. Figure 2 shows the test F1 for the *Separate* baseline which only considers in-domain training data compared with both a single-task domain adaptation model and a multi-task domain adaptation model. For simplicity, we only show the curve for the *Domain Mask* variant. As expected, we observe diminishing returns with additional in-domain training data on both tasks, but domain adaptation and multi-task domain adaptation methods suffer less from the diminishing return, especially on the NER task (Figure 2a). The curves for domain adaptation and multi-task domain adaptation also appear to be smoother, as they leverage more data to learn input representations, and thus are more robust.

When we have no in-domain training data, the problem reduces to unsupervised domain adaptation. Our framework applies here as well, and multi-task domain adaptation achieves performance close to the *Separate* baseline with only 200 in-domain training examples.

## 5.3 Model Variations

The multi-task domain adaptation framework is flexible regarding the number of domains and tasks, thus the number of datasets. Table 3 shows the results for several model variations, grouped by the number of training datasets. With one dataset, it is just the standard supervised learning setting, which reduces to our *Separate* baseline.

With two datasets, the framework can do multi-task learning (with two datasets from the same domain but different tasks), single task domain adaptation (with two datasets for the same task but from different domains), and a novel mismatch setting (with two datasets from *both* different domains *and* different tasks). As shown in the second section of Table 3, including additional training data – no matter from another task, domain or both – always improves the performance. A hidden factor not shown in the table is the additional dataset's size. For multi-task learning, since we are look at the social media domain, the additional dataset size is small. This is probably the reason why the *Mismatch* setting leveraging data from a different task *and* domain surprisingly outperformed multi-task learning. *Domain adaptation* enjoys both the benefits of a large amount of additional training data and an aligned task, thus achieving the best results among the two dataset settings.

| Dataset Numbers | Datasets / Methods | CWS | | | NER | | |
|---|---|---|---|---|---|---|---|
| | | Prec | Recall | F1 | Prec | Recall | F1 |
| One Dataset | Separate | 86.2 | 85.7 | 86.0 | 57.2 | 42.1 | 48.5 |
| Two Datasets | Multi-task | 87.7 | 86.2 | 86.9 | 59.1 | 44.9 | 51.1 |
| | Domain Adaptation | 88.7 | 87.1 | 87.9 | 68.2 | 48.6 | 56.8 |
| | Mismatch | 87.8 | 86.3 | 87.1 | 60.8 | 45.0 | 51.7 |
| Four Datasets | All Multi-task | 88.7 | 87.7 | 88.2 | 67.2 | 48.5 | 56.4 |
| | Multi-task DA | 89.7 | 88.3 | 89.0 | 60.2 | 52.3 | 59.9 |

Table 3: Model variations grouped by number of training datesets.

When conducting multi-task domain adaptation, we are leveraging four datasets. One concern is that the performance gains only come from additional training data, instead of the deliberately designed framework (Joshi et al., 2012). We thus also compare with a strategy which treats the same task for a different domain as a different task. The corresponding neural architecture is a shared BiLSTM with four separate task-specific models: we call it the *All Multi-task* setting. The results show that explicitly modeling data domains gives extra benefit than blindly throwing in more training data. We found the same benefits when experimenting with three datasets (instead of 2 or 4).

## 6 Related Work

The previous work on domain adaptation exclusively focused on building a unified model for *a* task across domain. However, we argue that a flexible framework for domain adaptation on several tasks simultaneously would be beneficial. To the best of our knowledge, the work that is closest to ours is Yang and Hospedales (2015), which provided a unified perspective for multi-task learning and multi-domain learning (a more general case of domain adaptation) under the same perspective of representation learning. However, they only focused on exploring the common ground of multi-task learning and multi-domain learning, and did not explore the possibility of having multi-task learning to help domain adaptation. We briefly review previous work on domain adaptation and multi-task learning below.

### 6.1 Domain Adaptation

In domain adaptation, or more general multi-domain learning, the goal is to learn a single model that can produce accurate predictions for multiple domains. An important characteristic of learning across domains is that each domain represents data drawn from a different distribution, yet share many commonalities. The larger the difference between these distributions, the larger the generalization error when learning across domains (Ben-David et al., 2010; Mansour et al., 2009).

As a result, a long line of work in multi-domain learning concerns learning shared representations, such as through identifying alignments between features (Blitzer et al., 2007, 2006), learning with deep networks (Glorot et al., 2011), using transfer component analysis (Pan et al., 2011), learning feature embeddings (Yang and Eisenstein, 2015) and kernel methods for learning low dimensional domain structures (Gong et al., 2012), among others. Another line sought for feature weight tying (Dredze and Crammer, 2008; Daumé III, 2007, 2009; Finkel and Manning, 2009; Kumar et al., 2010; Dredze et al., 2010) to transfer the learned feature weights across domains.

We combined the two lines and explored joint learning with multiple tasks.

### 6.2 Multi-task Learning

The goal of MTL (Caruana, 1997; Ando and Zhang, 2005) is to improve performance on different tasks by learning them jointly.

With recent progress in deep representation learning, new work considers MTL with neural networks in a general framework: learn a shared representations for all the tasks, and then a task specific predictor. The representations shared by tasks go from lower level word representations (Collobert and Weston, 2008; Collobert et al., 2011), to higher level contextual representations learned by Recurrent Neural Networks (RNNs) (Liu et al., 2016b; Yang et al., 2016; Peng et al., 2017) or other neural architectures (Liu et al., 2016a; Søgaard and Goldberg, 2016; Benton et al., 2017). MTL has helped in many NLP tasks, such as sequence tagging (Collobert et al., 2011; Peng and Dredze, 2016; Søgaard and Goldberg, 2016; Yang et al., 2016), text classification (Liu et al., 2016b,a), and discourse analysis (Liu et al., 2016c).

We expand the spectrum by exploring how multi-task learning can help domain adaptation.

## 7 Conclusion

We have presented a framework for multi-task domain adaptation, and instantiated a neural architecture for sequence tagging problems. The framework is composed of a shared representation learner for all datasets, a domain projection layer that learns one projection per domain, and a task-specific model layer that learns one set of feature weights per task. The proposed neural architecture can be trained end-to-end, and achieved the state-of-the-art results for Chinese word segmentation and NER on social media domain.

With this framework in mind, there are several interesting future directions to explore. First, we considered common domain adaptation schemas with our domain mask and linear projection. However, there are many more sophisticated methods that we can consider integrating into our model (Blitzer et al., 2007; Yang and Eisenstein, 2015). Second, we only experimented with sequence tagging problems. However, the proposed framework is generally applicable to other problems such as text classification, parsing, and machine translation. We plan to explore these applications in the future. Finally, our work draws on two traditions in multi-domain learning: parameter sharing (on the task specific models) and representation learning (the shared representation learner). We plan to explore how other domain adaptation methods can be realized in a deep architecture.